\documentclass[9pt,twocolumn,twoside]{osajnl}
\usepackage{array}
\usepackage{tabularx}
\usepackage{xltabular}
\usepackage{longtable}
\usepackage{wrapfig}
\usepackage{comment}
\usepackage{caption}
\usepackage{color,soul}
\usepackage{graphicx}

\journal{optica} % Choose journal (ao, aop, josaa, josab, ol, optica, pr)

\setboolean{shortarticle}{true}
\title{Machine Learning-aided Optical Performance Monitoring Techniques: An Overview}

\author{Dativa K. Tizikara}%, Jonathan Serugunda, and Andrew Katumba*}
\author{Jonathan Serugunda}
\author{Andrew Katumba}

\affil{Department of Electrical and Computer Engineering,School of Engineering, College of Design, Art and Technology, Makerere University, Kampala}

\affil[*]{Corresponding author: andrew.katumba@mak.ac.ug}

\begin{abstract}
Abstract
- Future communication systems are faced with increased demand for high capacity, dynamic bandwidth, reliability and heterogeneous traffic. To meet these requirements, networks have become more complex and thus require new design methods and monitoring techniques, as they evolve towards becoming autonomous. Machine learning has come to the forefront in recent years as a promising technology to aid in this evolution. Optical fiber communications can already provide the high capacity required for most applications, however, there is a need for increased  scalability and adaptability to changing user demands and link conditions. Accurate performance monitoring is  an integral part of this transformation.
In this paper, we review optical performance monitoring techniques where machine learning algorithms have been applied. Moreover, since many performance monitoring approaches in the optical domain depend on knowledge of the signal type, we also  review work for modulation format recognition and bitrate identification. We additionally briefly introduce a neuromorphic approach as an emerging technique that has only recently been applied to this domain.

\graphicspath{{./Images/}}

\end{abstract}

\setboolean{displaycopyright}{true}

\begin{document}

\maketitle

\section{Introduction}
Communication system evolution has led to the emergence of numerous novel applications with diverse capacity and reliability needs. As a result, many aspects of the network have had to become more complex and scalable. Fiber-optic channels currently meet the high capacity demands required, however, these optical networks will need to become elastic in the near future to support heterogeneous traffic and bitrates. Elasticity means that they are able to provide scalable bandwidth on demand and continuously adapt to ensure efficient resource utilization\cite{Liu2020} for example through the use of Bandwidth Variable Transcievers (BVT's) that can  generate variable bitrates \cite{gerstel2012}, Re-configurable Optical Add-Drop Multiplexers (ROADM's)\cite{gerstel2012,Berthold2008} that utilize wavelength selective switches to switch between flexible spectrum, and virtualization at network or transponder level\cite{Jinno2017}. A draw back of the resultant flexibility is that impairments experienced on the network vary with time because of the constantly changing light paths. To account for this, large safety design margins are often employed for such links to guarantee reliability which leads to inefficient use of network resources. In order guarantee good spectrum efficiency, low margins must be used hence it becomes crucial to monitor the performance of the optical links in real-time\cite{Morais2018,Liu2020}. In addition, networks are also becoming more intelligent which means that they have to be capable of self-optimization and self-diagnosing. In the case of fiber networks, this would mean that the network can detect anomalies along specific paths and therefore re-route traffic to other links, adapt the modulation format of signals based on link conditions and traffic and predict future network demands or failures along paths. In order to do this, they need to consistently acquire the quality of signals along the various paths.  
Optical performance monitoring (OPM) involves measuring and estimating different physical parameters of transmitted signals and components in an optical network either at the receiver or at an intermediate node along  the path \cite{Dong2016}. This enables the transmission system parameters relating to the channel quality to be known so that they can be compensated for. Common parameters include Chromatic Dispersion (CD), Polarization Mode Dispersion (PMD), Optical Signal to Noise Ratio (OSNR), Q-factor, Polarization Dependent Loss (PDL) and fiber non-linearities. Conventional OPM techniques required complete recovery of the transmitted signal to be able to deduce these parameters. However, so as to compensate for resultant signal degradation, these performance metrics need to be known at distributed points on the fiber link hence traditional techniques would add significant complexity and cost to the monitoring system which is not desired. Machine Learning(ML) has emerged as a key technique that  can be used to process the received signal at different points and learn relationships between different characteristics of the received signal  and impairments without having to completely demodulate the signal\cite{Dong2016,Khan2019}. In order to reduce costs, it is also required to monitor multiple impairments simultaneously and independently.

This paper aims to survey existing work where machine learning has been applied to aid in OPM and discuss the performance of the different techniques. Moreover, since the bulk of the techniques employed in the current literature require advance knowledge of the signal type, we also review some works that identify the modulation format and bitrate. Furthermore, we briefly explore work on photonic reservoir computing which has more recently been shown to be applicable to modulation format recognition. 
\section{Related Work}
There are a number of review works on utilization of Machine Learning for  various applications in optical networks. Existing and future technologies for OPM for both direct and coherent detection systems are reviewed in \cite{Dong2016}, however, their work presented a broad range of techniques and did not focus on ML techniques. A detailed review of the different optical ML techniques was given in \cite{Khan2019a,Musumeci2019,Mata2018a} highlighting how they have been used in optical communications and networking functions such as for OPM, fault detection, non-linearity compensation and software defined networking. They, however,   had limited coverage of OPM and Modulation Format Recognition (MFR). A detailed survey on OPM and MFR has been done in \cite{Saif2020}. 
We update the current literature in this work as well as include the application of photonic reservoir computing which has only recently been applied to modulation format identification. The work in  \cite{Amirabadi2019} considered a detailed description of machine learning techniques and reviewed works that had applied them in the optical communications space.
%\label{sec:examples}
%\label{sec:examples}
%\section{Summary of Machine Learning Techniques used in Optical Communications}

\section{Feature Selection for OPM}

Machine learning algorithms typically take input data features and learn relationships between them, thereby being able to group the inputs in a certain way or map the relationship to a function that can predict a required output. 
For OPM, the outputs are the type of impairment and its amount, while the inputs are signal representations. The signal representations are obtained from monitoring the signal waveform, polarization or spectrum \cite{Dong2016} or from Digital Signal Processing (DSP) techniques in the electrical domain after detection in direct detection schemes. Coherent receivers already include powerful DSP blocks and input features can directly be obtained from the asynchronously sampled output of these blocks \cite{Tanimura2016,Cho2019}, or from constellation diagrams that can be constructed from them \cite{Kashi2017,Wang2017}. 

The output of these various methods can then be utilized in the form of direct images or their properties, or statistical representations for example histograms, means, variances and moments to extract different features that can then be fed to the machine learning processing blocks. The features are chosen either manually by visual inspection or learnt by the ML algorithm and they show a clear distinction among different types of impairments and their levels. Table \ref{table:1} shows a summary of monitored impairments for different feature types in current works.
\begin{table*}
 \centering
  \captionsetup[]{justification=centering}
  \caption{\bf Summary of features and monitored impairments used in current works}
\begin{tabular}{llllll}
\hline
Feature Source & OSNR & PMD* & CD & Non-linearity &Crosstalk \\ \hline
Eye diagram&\checkmark\cite{Thrane2017,Jargon2009a,Wu2009a}&\checkmark\cite{Jargon2009a,Wu2009a,Skoog2006b}&\checkmark\cite{Wu2009a,Skoog2006b}&\checkmark\cite{Wu2009a}&\checkmark\cite{Skoog2006b}\\
ADTP (Phase portrait)&\checkmark\cite{Jargon2009,Tan2014,Wu2011,Fan2018}&\checkmark\cite{Jargon2009,Anderson2009,Tan2014,Wu2011,Fan2018}&\checkmark\cite{Jargon2009,Anderson2009,Tan2014,Wu2011,Fan2018}&&\checkmark\cite{dods2006,Celik2018}\\
Asynchronous sampled signal amplitude&\checkmark\cite{Khan2012}&\checkmark\cite{Khan2012}&\checkmark\cite{Khan2012}\\
Asynchronous Constellation diagram&\checkmark\cite{Jargon2010}&\checkmark\cite{Jargon2010}&\checkmark\cite{Jargon2010}\\
Spectrum&\checkmark\cite{Wang2019b}\\
AAH&\checkmark\cite{Wan2018a,Cheng2020,Khan2017,Xia2019}\\
Asynchronous eye diagram&\checkmark\cite{Vitor2012}&\checkmark\cite{Vitor2012}&\checkmark\cite{Vitor2012}\\
Optical power&\checkmark\cite{Zheng2020}\\
Asynchronous sampled raw data&\checkmark\cite{Tanimura2016,Cho2019,Wang2019} \\
Constellation diagram&\checkmark\cite{Kashi2017,Wang2017}&&&\checkmark\cite{Caballero2018}\\
ASCS&\checkmark\cite{Fan2019}&\checkmark\cite{Fan2019}&\checkmark\cite{Fan2019}\\
IQH&\checkmark\cite{Saif2021}&&\checkmark\cite{Saif2021}&&\\
Stokes-space constellation&\checkmark\cite{Xiang2021}&&\\

\hline
PMD* refers to 1st order PMD in this paper
\end{tabular}
 \label{table:1}
\end{table*}
\subsection{Eye Diagrams}

An eye diagram is a graphical representation of a signal waveform showing the amplitude distribution over one or more bit periods, with the symbols overlapping each other. The quality of the signal can then be determined from various characteristics of the eye opening for example jitter, SNR, dispersion, non-linearities. 

\begin{figure}[ht]
\centering
\fbox{\includegraphics[width=\linewidth]{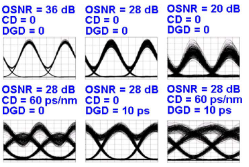}}
\caption{Impact of various impairments on the eye diagram of an RZ signal\cite{Wu2009a}.}
\label{fig:eyediagram}
\end{figure}

Eye diagrams have been used in the current literature to monitor OSNR, PMD, CD, non-linearity and crosstalk. Figure \ref{fig:eyediagram} shows the eye diagrams for an RZ signal subjected to different impairments\cite{Wu2009a}. 
Visual inspection shows that different impairments and different levels of the same impairment produce distinct characteristics.These characteristics can be exploited by applying image processing techniques such as in  \cite{Skoog2006b}, by defining statistical features from the sampled amplitudes for example means and variances at specific points on the eye diagram \cite{Thrane2017}, or by calculating the widely used parameters of the eye diagrams \cite{Jargon2009a,Wu2009a}.
Construction of eye diagrams is dependent on the modulation format and requires timing synchronisation hence some form of clock recovery is required which can be expensive. An eye diagram also has no phase information about the signal.
\subsection{Asynchronous Delay Tap Plots (ADTP's)}
\begin{figure}[ht]
\centering
\fbox{\includegraphics[width=\linewidth]{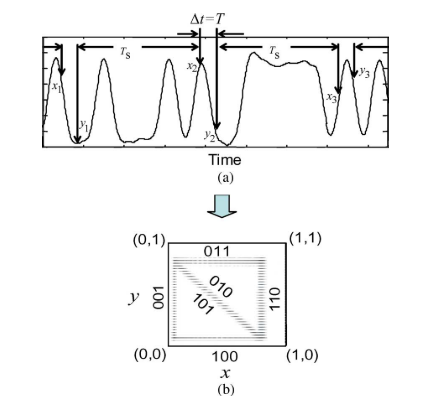}}
\caption{Generation of phase portrait\cite{Anderson2009}}
\label{fig:phaseportrait}
\end{figure}

This technique also provides a visual representation of a signal known as a phase portrait.  The signal waveform is split and one part of the signal delayed by a certain amount $\Delta$\text{t}. The signal and its delayed version are then sampled at the same instant and the pair of values (x,y) obtained plotted in a 2D histogram \cite{dods2006,Celik2018}. Figure \ref{fig:phaseportrait} illustrates how a phase portrait is created from delay tap sample pairs \cite{Anderson2009}.The sampling period, $T_s$  is independent of the bit duration, $T$ and can therefore be several magnitudes larger. The portraits can be treated as images and exploited using pattern recognition \cite{Anderson2009,Tan2014,Anderson2009a} and then image processing algorithms applied, or specific features extracted from them  for example the work in \cite{Jargon2009} divided the phase portrait into quadrants and then defined statistical means and standard deviations of the (x,y) pairs and radial coordinates.

\begin{figure}[ht]
\centering
\fbox{\includegraphics[width=\linewidth]{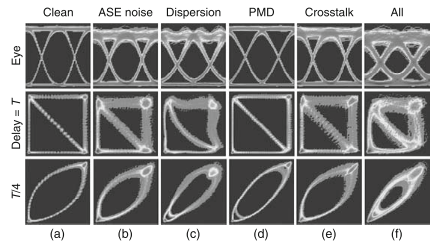}}
\caption{ADTP's for a 10 Gb/s NRZ signal in the following scenarios: a) OSNR = 35 dB, b) OSNR = 25 dB, c) CD = 800 ps/nm , d) DGD = 40 ps, e) crosstalk = -25 dB and f) a combination of (b-f) \cite{Celik2018}}.
\label{fig:adtpcompare}
\end{figure}

Phase portraits are also dependent on the signal properties such as bitrate and modulation format and the tap delay. The tap delay is usually a certain fraction or multiple of the symbol rate and thus needs to be adjusted exactly for different datarates to allow accurate monitoring \cite{Khan2011}. ADTP's have been used for multiple impairment monitoring of OSNR, CD, crosstalk and 1st order PMD.In figure \ref{fig:adtpcompare}, the effect of various impairments on the ADTP of a $\text{10 Gbps}$ NRZ signal at two different delays, $T$ and $T/4$, as well as the corresponding eye diagrams are shown. \cite{Celik2018}.

\subsection{Asynchronous Amplitude Histograms(AAH's)}
AAH's are obtained from random asynchronous sampling of the signal within the bit period. The authors in \cite{Chen2004} showed that with a sufficient number of samples, the amplitude distribution can be accurately represented within a bit period. The amplitude samples are arranged in bins corresponding to their level, and then the count of samples within each bin is plotted against the bin. This is in contrast to the synchronous AH where the considered samples are within a specific window for example 10\% \cite{Celik2018} of the bit period around the center of the eye diagram at the optimal decision time. The peaks in the AAH correspond to the samples around the maximum and minimum values of the eye, and the samples in between correspond to those around the crossings of the rising and falling edges of the waveform.

\begin{figure}[ht]
\centering
\fbox{\includegraphics[width=\linewidth]{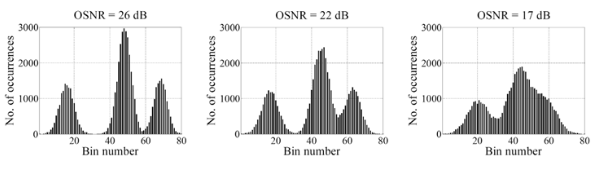}}
\caption{Impact of varying OSNR on the AH of a 16-QAM signal\cite{Khan2017b}}.
\label{fig:aah}
\end{figure}

Amplitude histograms are simple and transparent to the transmitted signal characteristics such as modulation format and bitrate, however, the contribution of each individual impairment cannot be independently extracted hence they have not been used for multiple impairment monitoring. Furthermore, the monitoring accuracy is dependent on the number of samples \cite{Wan2018,Dong2016,Cheng2020}. The count of occurrences at each bin can then be used as input features such as in \cite{Wan2018,Khan2017b}. \cite{Xia2019} additionally used the variance of the amplitude values in each bin. 
Figure \ref{fig:aah} shows results of varying the OSNR on the AAH for a 16-QAM signal\cite{Khan2017b}.
\subsection{Asynchronous Single Channel Sampling (ASCS)}
In this method, the signal $s(t)$ is sampled asynchronously  using one tap, and then the samples are shifted by k samples and the sample pairs $s_i(t)$ and $s_{i_{+_k(t)}}$ used to construct a phase portrait. This method is less expensive than two-tap sampling \cite{Yu2014,Fan2019,Fan2020}.The generated phase portraits can be used as images for example in \cite{Fan2019,Fan2020}.
\begin{figure}[ht]
\centering
\fbox{\includegraphics[width=\linewidth]{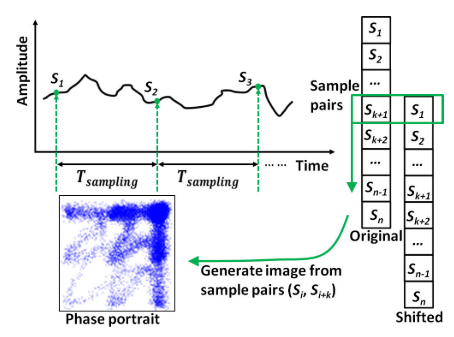}}
\caption{Generation of a phase portrait from ASCS \cite{Fan2020}}.
\label{fig:ascs}
\end{figure}
\subsection{Constellation Diagrams}
A constellation diagram is a graphical representation of a digitally modulated signal, where received samples are represented in an I/Q diagram.They are used in coherent detection schemes and can be generated by techniques such as linear optical sampling \cite{Dorrer2005}, however, since coherent receivers already have embedded DSP blocks, they can be directly constructed from the asynchronously sampled data output of the DSP. Thereafter, they can be used to generate manually defined features for example \cite{Caballero2018} defined tangential and normal components of the noise of each symbol and then used averages and amplitude noise covariances as inputs, or their images can be directly input the ML algorithm for image processing without the need for manual feature generation for example in  \cite{Wang2017}.

Constellation diagrams have only been used to measure OSNR and non-linear noise in coherent detection system since the coherent receiver can already compensate for CD and PMD and therefore these impairments can be directly monitored.
\subsection{In-phase Quardrature Histograms}
IQH's were proposed in \cite{Saif2019z} as an extension of AAH's to include phase information for coherent systems. They contain similar information as constellation diagrams but with an additional representation of the amplitude in colour. They showed that it can be used to identify OSNR, PMD and CD although performance degraded in the presence of multiple impairments. 
\begin{figure}[ht]
\centering
\fbox{\includegraphics[width=\linewidth]{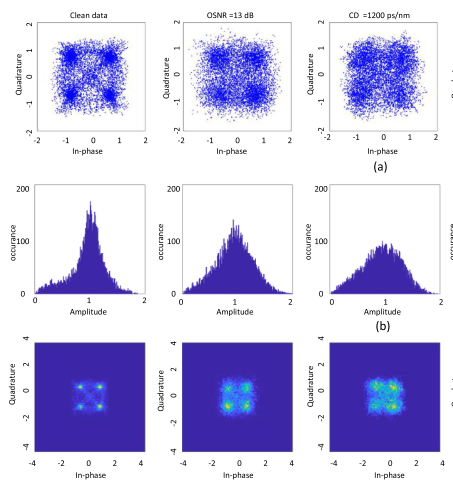}}
\caption{Comparison of constellation diagram, AAH and IQH \cite{Saif2019z}.}
\label{fig:IQH}
\end{figure}
\cite{Saif2021} derived 1D features from projections of IQH's on diagonal and horizontal axes. 
\subsection{Stokes Space Constellation}
\begin{figure}[ht]
\centering
\fbox{\includegraphics[width=\linewidth]{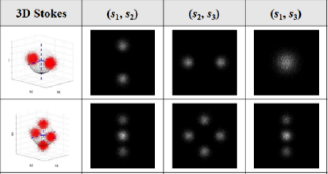}}
\caption{3D Stokes constellation of a BPSK and QPSK signal, as well as their corresponding projections in the 2D Stokes planes at OSNR=18dB \cite{Zhang2020b}.}
\label{fig:stokes}
\end{figure}

This diagram is obtained by plotting the last three components of the Stokes vector of the received complex signals from a coherent receiver in a 3D Stokes space. Different modulation formats present a specific number of distinguishable clusters in this space. \cite{Boada2015,Mai2017,Szafraniec2010}. The authors in \cite{Xiang2021} obtained the cumulative distribution function (CDF) of one Stokes parameter while \cite{Zhang2020b} projected the constellation onto three different 2D planes and used the resultant plots as images such as in figure \ref{fig:stokes}.
\subsection{Other methods}
The nature of asynchronous sampling means that certain information in the signal is lost, which could make it difficult in some cases to separate the effects of different impairments from the overall received signal in case they produce similar changes in the plots \cite{dods2006}. Furthermore, there is overlap in the distribution of signal amplitudes which makes it more challenging to extract individual distributions from AAH's in practice \cite{Khan2011}. Asynchronous eye diagrams \cite{Costa2012} and asynchronous constellation diagrams \cite{Jargon2010} can be constructed to mitigate this. In addition, \cite{Khan2011} also proposed asynchronously sampled amplitudes as a solution for better CD monitoring since previous works had shown that CD was severely impacted by changes in OSNR and Differential Group Delay and to eliminate the requirement for continuously adjusting the tap delay for multiple bitrates.

Optical spectral data from an optical spectrum analyzer (OSA) and optical power have also been used in \cite{Wang2006} and  \cite{Zheng2020} respectively.

%\label{sec:examples}
\section{Survey of ML-based Optical Performance Monitoring techniques}
\subsection{OPM for networks using Direct Detection}
OPM modules in systems employing direct detection can be as straightforward as a photo-detector in combination with an Analog to Digital Converter.

In \cite{Skoog2006b}, multiple Support Vector Machines (SVM's) were utilized to classify different impairments using images of eye diagrams, characterised by 23 low order zernike moments. Simulation data was used to train the model after impairments of CD, PMD and cross talk were applied. Four SVM's were required, one for each impairment and an additional one for the normal case since they are binary classifiers. The number of input images used for training the model were 31, 107, 20 and 6 for CD, PMD, crosstalk and normal respectively. Experimental verification was then done using the model. Results collected from 3, 11 and 3 images for CD, PMD and crosstalk respectively  showed that the method could classify the simulated and experimental data with accuracies of 95\% and 60\%.  However, it could only identify the type of impairment but not the amount. Application of a nearest neighbors technique after the SVM was proposed to enable this.

In \cite{Jargon2009a}, an Artificial Neural Network (ANN) connsisting of a single hidden layer and 12 hidden neurons was demonstrated to predict multiple impairment levels simultaneously. The eye diagrams of signals with different bitrates and modulation formats i.e. 10 Gb/s non-return to zero on-off keying  (NRZ-OOK) and 40 Gb/s return-to-zero differential phase shift keying (RZ-DPSK), to which different combinations of CD, PMD and OSNR had been applied, were used to train the ANN. 4 input features were extracted from each of 189 eye diagrams i.e. (Q-factor, closure, jitter and crossing amplitude/level of transition between adjacent zeros for NRZ-OOK/RZ-DPSK respectively). 125 eye diagrams were used for training while 64 were used for validation. The ranges used for OSNR, CD and Differential Group Delay (DGD) were 16-32 dB, 0-800 ps/nm  and 0-40 ps for NRZ-OOK, and 16-32 dB, 0-60 ps/nm and 0-10 ps for RZ-DPSK. A correlation coefficient of 0.91 was achieved for the NRZ-OOK signals while 0.96 was found for the RZ-DPSK case. A similar investigation was done in later work in \cite{Jargon2009}, but using 7 manually defined parameters from ADTP's as input to a single layer, 28 neuron ANN for the 10 Gbps NRZ-OOK case. A higher correlation coefficient of 0.97 was obtained over similar impairment ranges. The work was further extended to monitor the same three impairments for a 40 Gbps RZ-QPSK signal and manually defined input parameters using asynchronous constellation diagrams \cite{Jargon2010}. An identical ANN was used in \cite{Jargon2009} achieving a correlation coefficient of 0.987, and root mean square errors (RMSE's) of 0.77 dB, 18.71 ps/nm and 1.17 ps for OSNR, CD and DGD respectively. The impairment ranges tested were 12-32 dB, 200 ps/nm and 0-20 ps for OSNR, CD and DGD.
The same ANN technique with 1 hidden layer and 12 hidden neurons was used in \cite{Wu2009a} to monitor the effect of multiple impairments on 40 Gbps RZ-OOK and RZ-DPSK data signals. 4 input parameters (Q-factor,eye-closure, RMS jitter and RMS jitter) were defined from eye diagrams. The ANN was trained and tested with data from both simulation and experiment. In the simulation, 125 and 64 eye diagrams were used for training and validation respectively, achieving a correlation coefficient of 0.97 and 0.96 for OOK and DPSK respectively and average errors for OSNR, CD and DGD of 0.57 dB, 4.68 ps/nm and 1.53 ps for OOK and 0.77dB, 4.47 ps/nm and 0.92 ps for DPSK. The simulations were followed up with an experiment, in which 20 and 12 eye diagrams were used for training and testing respectively to estimate OSNR and CD.The results showed a better performance than simulation with 0.99 correlation coefficient for both signals. The average errors  for OOK were 0.58 dB and 2.53 ps/nm while those for DPSK were 1.85dB and 3.18 ps/nm. The ranges tested for OSNR, CD and DGD were 16-32 dB, 0-60 ps/nm and 1.25-8.75 ps. The authors then monitored the impact of  accumulated fiber non- linearity in a 40 Gb/s RZ-DPSK Wavelength Division Multiplexed (WDM) system consisting of  3-channels using a simulation in which additional features consisting of statistics of the 1 and 0 values were defined giving a total of 8 inputs. The input optical power was varied from -5 to 3 dBm, while OSNR, CD and DGD were tested over the ranges from 20-36 dB, 0-40 ps/nm and 0-8 ps. Equally good results were obtained: correlation coefficient of 0.97, and mean error of 0.46 dB, 1.45 dB, 3.98 ps/nm and 0.65 ps for optical power, OSNR, CD and DGD from 135 training samples and 32 testing samples.

In \cite{Anderson2009} CD and DGD were simultaneously measured for a 40 Gb/s NRZ-DPSK signal. ADTP's were generated and then kernel based ridge regression applied to predict the impairments using 900 features. Simulation was done for various combinations of CD, DGD and OSNR ranging from 0-700 ps/nm, 0-20 ps and 13-26 dB respectively. 1200 phase portraits consisting of 900 features each were used for training, and independent training for a single impairment in the presence of all other impurity ranges was done. 500 phase portraits were used for verification. RMSE's of +/-11ps/nm and +/-0.75 ps for CD and DGD respectively were achieved. Experimental verification was also done using a split of 1500:500 phase portraits for training:validation for OSNR, CD and DGD ranging from 15-25 dB, -400 to 400 ps/nm and 0-22.5 ps. The total training time was 3hrs and RMSE of +/-11ps/nm and +/-1.9ps for CD and DGD obtained. Prior knowledge of modulation and bit rate was assumed.

OSNR, PMD and the magnitude and sign of CD were monitored in \cite{Khan2012} using an ANN whose input features were derived from the empirical moments of amplitude samples. The ANN consisted of a single hidden layer with 42 neurons and was trained with simulation data for 56 Gb/s RZ-DQPSK and 40 Gb/s RZ-DQPSK and DPSK signals. For each datarate-modulation format combination, 3627 groups of moments were collected over varying ranges of OSNR(10-26 dB), CD (-500 to 500 ps/nm) and DGD (0-14 ps). A root mean square errors of 0.1 dB was obtained for OSNR in all three cases while the values obtained for CD and DGD were CD(27.3, 29, 17  ps/nm)  and DGD (0.94, 1.3, 1 ps) for 40 Gb/s RZ-DQPSK, 56 Gb/s RZ-DQPSK and 40 Gb/s RZ-DPSK systems respectively. The authors proposed increasing the number of moments to improve the results.\newline
Table \ref{table:2} summarizes existing works for direct detection systems.

\begin{table*}
 \centering
  \captionsetup[]{justification=centering}
  \caption{\bf Summary of existing OPM works for direct detection}
\begin{tabular}{p{4cm}p{5cm}p{5cm}p{1cm}}
\hline
\textbf{ML Algorithm}&\textbf{Signal Type} &\textbf{Impairment (range)}&\textbf{Ref}\\
\hline
SVM&&CD, DGD, Crosstalk &\cite{Skoog2006b}\\
ANN(1;12) & 10 Gbps NRZ-OOK&OSNR(16-32),   CD(0-800), DGD(0-40) & \cite{Jargon2009a}\\
&40 Gbps RZ-DPSK&OSNR(16-32), CD(0-60), DGD(0-10)\\
ANN(1;28)&10 Gbps NRZ-OOK&OSNR(16-32), CD(0-60),DGD(0-10)&\cite{Jargon2009}\\
ANN(1;28)&40 Gbps RZ-QPSK&OSNR(12-32), CD(0-200), DGD(0-20)& \cite{Jargon2010}\\
ANN(1;12)&40 Gbps RZ-OOK&OSNR(16-32), CD(0-60), PMD (1.25-8.78) &\cite{Wu2009a}\\
&40 Gbps RZ-DPSK&OSNR(16-32),CD(0-60),PMD (1.25-8.78)  \\  
&40 Gbps RZ-DPSK, 3 channel WDM & optical power(-5 to 3 dBm), OSNR(20-36), CD(0-40) PMD(0-8)\\
ANN(1;40)&40 Gbps QPSK &OSNR(10-30), CD(0-200), DGD(0-25) &\cite{Costa2012}\\
ANN(1;3) &32 Gbd PDM 64-QAM& OSNR(4-30)&\cite{Thrane2017}\\
ANN(1;42)&40 Gbps RZ-DQPSK&OSNR(10-26),  CD(-500-500), DGD(0-14)&\cite{Khan2012}\\
&56 GbpsRZ-DQPSK&\\
&40 Gbps RZ-DPSK\\
Kernel ridge regression &40 Gbps NRZ-DPSK&CD(0-700), DGD(0-20)&\cite{Anderson2009}\\
&&CD(-400 to 400)\textsuperscript{*}, DGD(0-22.5)\textsuperscript{*}& \\
PCA&10/20 Gbps RZ-OOK, 40/100 Gbps PDM RZ-QPSK, 100/200 Gbps PDM NRZ 16-QAM &OSNR(14-28), CD(-500-500), DGD(0-10)&\cite{Tan2014}\\
ANN(1;12)&100 Gbps QPSK&OSNR(14-32),CD(0-50),DGD (0-10)&\cite{Wu2011}\\
MTL-ANN&28 Gbd NRZ-OOK, PAM4, PAM8 &OSNR(10-25), (15-30), (20-35)& \cite{Wan2018}\\
MT-DNN(4:100,50,30,2) with transfer learning &10 Gbd PDM 16 , 64 QAM & OSNR(14-24), (23-34)&\cite{Cheng2020}\\
MTL-ANN&10 Gbd NRZ-QPSK,32 Gbd PDM-16QAM&OSNR(1-30)&\cite{Zheng2020}\\
MTL-CNN&10/20 Gb/s RZ-OOK, NRZ-OOK, NRZ-DPSK &OSNR(10-28), CD(0-450), PMD(0-10)&\cite{Fan2018}\\
MTL-CNN&60/100 Gbps QPSK, 16, 64 QAM&OSNR(10-28), CD(0-450), PMD(0-10)&\cite{Fan2019}\\
MTL-DNN&14/28 GBd QPSK, 16QAM &OSNR (10-24),(15-29), CD (0, 858.5, 1507.9)&\cite{Luo2021}\\
ANN(5;40)&4,16,32,64,128 QAM&OSNR(15-20)&\cite{Zhang2018}\\
\hline
\multicolumn{4}{p{14cm}}{All units for OSNR,CD,PMD are in dB,ps/nm and ps respectively. *indicates experimental results and simulation results otherwise. ANN(x;y): x and y represent the number of hidden layers and neurons in the hidden layer respectively.}
\end{tabular}
 \label{table:2}
\end{table*}

The work presented in \cite{Tan2014} monitored multiple impairments and identified both modulation format and bit rate using Principal Component Analysis (PCA). Input features were derived from images of ADTP's and the method was shown to be suitable for heterogeneous networks. Simulations were used to generate 26,208 ADTP's from different combinations of impairments, modulation schemes and bitrate. Previous methods seen so far have assumed knowledge of both bitrate and modulation. The impairments were varied in the range 14-28 dB (OSNR), -500 to 500 ps/nm (CD) and 0-10 ps (DGD). The signal combinations used were 10 and 20 Gb/s RZ-OOK, 40 and 100 Gb/s PDM RZ-QPSK and 100/200 Gb/s PDM NRZ 16-QAM. The results showed an overall mean estimation error of 1 dB (OSNR), 4 ps/nm (CD) and 1.6 ps (DGD). The performance of the method under fiber non linearity was also investigated and found to be slightly less accurate, increasing the mean errors to 1.2 dB for OSNR, 12 ps/nm for CD and 2.1 ps for DGD. To mitigate this, selection of additional features to characterise different non-linearity coefficients and link/span lengths was proposed. In this way CD, OSNR and DGD could be monitored without advance knowledge of the signal type, it was part of the training data.

The authors in \cite{Thrane2017} used an ANN  for in-band OSNR monitoring on 32 Gbaud directly detected PDM-QAM signals. The input features were selected from eye diagrams. In addition to the modulation format, this method required knowledge of the pulse shape therefore it was necessary to train a separate neural network for each pulse-MF pair. It was composed of 1 hidden layer, 3 hidden neurons and only one input feature i.e. the variance at the maximum amplitude points on the eye diagram. Experimental verification was done for OSNR's in the range of 4-30 dB but only in white Gaussian noise. The results showed that OSNR estimation was accurate between 4-17 dB with a mean error of 0.2 dB but worsened from 17-30 dB. This was attributed to the fact that eye diagrams at higher OSNR's did not vary very significantly and hence had less distinguishable features. Since real transmission channels face other impairments, simulation was done for chromatic dispersion (CD) and the method found to be unimpaired up to 250 km on a dispersion uncompensated link. Verification of the method in the presence of other effects was left to future work.

Multi- impairment monitoring was investigated in \cite{Wu2011} using a single layer, 12 hidden neuron ANN that was trained with simulated data from 180 ADTP's. 7 statistical features were extracted from each ADTP obtained from sampling a 100 Gb/s QPSK signal over impairment ranges of OSNR (14-32 dB), CD (0-50 ps/nm) and DGD (0-10 ps) respectively. The validation was done with 144 samples. Balanced detection was shown to perform better than single ended detection through simulation with correlation coefficients of 0.995 and 0.96 respectively. The RMSE's were obtained as OSNR (1.62/0.45 dB), CD (8.75/3.67 ps/nm) and DGD (7.02/0.8 ps) for single/balanced detection. Experimental data was used to validate the performance for balanced detection and produced correlation of 0.997.

Simultaneous monitoring of PMD, CD and OSNR using a single layer, 40 neuron ANN was shown in \cite{Costa2012} used a single hidden layer ANN with 40 neurons. Parametric asynchronous eye diagrams (PAED's) of a 40 Gbps QPSK signal were collected, from which 24 statistical position features were extracted. In this work, RMSE's of <20 ps/nm , < 1.3 ps and 1.5-2 dB were found via simulation for impairments in the ranges 0-200 ps/nm, 0-25 ps and 0-30 dB. 

In \cite{Wan2018} a Multi Task Learning ANN (MTL-ANN) was investigated using features extracted from amplitude histograms and used to acquire both OSNR and the modulation format. Simulations were done on 28-Gbaud NRZ-OOK, PAM4 and PAM8 over an OSNR range of 10-25 dB, 15-30 dB and 20-35 dB respectively and CD range of -100 to 100 ps/nm. A total 9072 and 1008 simulated AH's were used for training and testing respectively. Different combinations of OSNR and modulation format at specific CD values were tested achieving a MSE of 0.12dB. Experimental verification was done  for OSNR ranges of 14-29, 17-32 and 22-37 dB for OOK, PAM, PAM8 and datasets consisting of 4320 and 480 AH's for training  and testing. The results showed higher accuracy than single task learning ANNS (STL-ANN's), achieving MSE of 0.11 dB compared to 0.4 dB for a STL-ANN with a similar structure. This method required optimization of the bin number. Fewer bins were shown to have less accuracy while more bins led to a more complex ML structure. The authors used an optimal number of 100 in this work. 

OSNR and modulation format monitoring was done in \cite{Cheng2020} using a mutli-task deep neural network with transfer learning (DNN-TL) using AH's as inputs. The DNN was trained with 400 AH's generated from simulation and then experimental verification for PDM-16 and 64-QAM , 10 Gbaud signals was done and the results achieved  RMSE of 1.09 dB for OSNR ranging between 14 to 24 dB for PDM-16 QAM and 23 to 34 dB for 64-QAM respectively. The ANN structure had 4 hidden layers with 100/50/30/2 neurons respectively. Application of transfer learning was able to reduce the required training samples from 322 to 243 (243) for the same RMSE. 

A modulation format independent method was proposed to monitor the OSNR for a WDM system in \cite{Zheng2020}. Optical power measured at different center wavelengths was used as input features to a MTL-ANN with 64 neurons per layer. 5 samples for each OSNR (1-30 dB) were collected and a ratio of 70:30 samples was used for training: testing and shown experimentally  to estimate the OSNR with a Mean Absolute Error (MAE) of  0.28 dB and RMSE of 0.48 dB for both the 10 Gbaud NRZ-QPSK and 32 baud PDM-16QAM over an OSNR range of 1-30 dB. It was also shown to be insensitive to CD and PMD.  The same ANN was shown to be capable of simultaneously monitoring baud rate and launch power without deploying two additional ANN's. For launch power in the range of 0-8 dBm, MAE and RMSE were 0.034 dB and 0.066 dB respectively.

A MTL-CNN was used in \cite{Fan2018} to do multiple impairment monitoring in combination with joint bit rate and modulation format identification. 6600 Phase portraits were generated from simulations of six different signal types i.e. 10/20 Gb/s RZ-OOK, NRZ OOK and NRZ-DPSK and impairments varied over the ranges 10-28 dB, 0-10 ps and 0-450 ps/nm for OSNR, DGD and CD respectively. 90\% of the images were used to train the CNN while 10\% were reserved for testing. The results showed RMSE's of 0.73 dB, 1.34 ps/nm and 0.47 ps. The same authors improved their method by using phase portraits from ASCS in \cite{Fan2019} and features from the various CNN layers as opposed to only those in the last layer. In this method, the features were extracted from all the layers and transformed into the same space and then multiple tasks were trained for each of OPM, MFR and bitrate identification (BRI). 60/100 Gb/s signals for 3 modulation formats  QPSK, 16 and 64-QAM were generated by simulation and the same impairment ranges and number of phase portraits were used. RMS errors of  1.52 ps/nm, 0.81 dB, and 0.32 ps were obtained.

In the work presented in \cite{Luo2017}, adaptive ADTP's and AAH's were used as multiple inputs to a multi-task DNN to monitor OSNR in the range 10-24 dB and 15-29 dB for QPSK and 16 QAM signals respectively and identify the bitrate, modulation format and chromatic dispersion. 2 baudrates (14/28) and 3 values of CD (0, 858.5 and 1507.9 ps/nm) were experimentally tested. In the AADTP, a single ADC is used to sample the data generating $x_m$, samples ($m>=1$) as opposed to two tap delay sampling and then a fixed time delay is introduced by setting the second sample pair as a $y_m=x_m+n$ , $n>=1$. The same samples are used to generate AAH's. 36,000 AADTP's and AAH's were generated and 28,800 of them used to train the DNN. The method achieved a MAE of 0.2867 dB and CD identification accuracy of 99.83\%.

A simple 3 layer ANN was used in \cite{Zhang2018} to jointly monitor OSNR (15-20 dB) and identify the MF in an IM-DD QAM-OFDM system. Two ANN's were used; one for MFI and then once the MF was known, passed to the second ANN which was trained for each modulation format separately to identify the OSNR. AAH's were derived from the IQ output by considering either the I or Q samples of 4,16, 32, 64 and 128 QAM signals. To improve the OSNR accuracy at low OSNR's, 5 distinct features were calculated from the AH's and used as input to the second ANN i.e. mean, variance, range, interquartile and median. The errors obtained OSNR prediction were < 1dB. 

Table \ref{tab:3} shows the performance of the different techniques that have been surveyed.
\subsection{Coherent Detection}
Coherent detectors already incorporate  impairment compensation techniques at the receiver and therefore linear impairments - CD and PMD  can already be monitored. OSNR then becomes the key impairment that still requires monitoring. Many of the previous methods discussed required the careful selection of features from sampled data. These features varied for different system parameters. As networks evolve, they will transmit data at varying bitrates and modulation formats which may change randomly hence more advanced techniques are required.

The authors in \cite{Tanimura2016} used experimental data to train a Deep Neural Network (DNN) to monitor OSNR of a 16 GBd DP-QPSK signal with asynchronously sampled raw data from a coherent receiver. The DNN was trained with 3 different hidden layer structures (1,3,5) each comprising  512 neurons, and 3 training sample sizes (4000, 40,000 and 400,000). The 4 tributary output from the coherent receiver was then fed to the DNN, each tributary containing 512 samples generated from experiment. The 5 layer, 400,000 case was selected as the best case. The trained DNN was then used to test 10,000 samples resulting in an average error of 1.6 dB over an OSNR range of 7.5 - 31 dB. 
 
In \cite{Cho2019}, the same method was extended to  a Convolutional Neural Network (CNN) which was trained with experimental data containing 1,000,000 samples from 14 GBd and 16 GBd DP-QPSK, 16-QAM and 64-QAM signals that had been subjected to different OSNR's within a range of 11-33 dB. The CNN was validated using 10,000 test samples for each modulation format. The results obtained showed a bias error of less than 0.3 dB, however the training phase took several hours. They showed the method to be insensitive to CD and left non-linearity to future work.

A single layer ANN with 6 hidden neurons was used in \cite{Kashi2017} to estimate  non-linear noise present in a 56.8 GBd DP 16-QAM signal transmitted over fiber channels with varying characteristics for example transmission distances, optical power, number of channels , types of fiber etc. The ANN was provided with the link parameters as well as amplitude noise co-variance (ANC) of the input symbols resulting from fiber non-linearity for 240 simulated cases. 70\% of the samples were used for training and 30\% for testing resulting in the errors obtained in the OSNR being less than 0.6 dB for two experimental cases.

In \cite{Caballero2018a} a neural network was used to estimate both linear and non-linear noise simultaneously using input features derived from constellation plots  and the amplitude noise co-variance. The ANN consisted of 1 hidden layer and 7 neurons and was trained with a 35 Gbd DP-16 QAM signal transmitted over different WDM channels, with varying fiber types and lengths of 320-1200 km, launch power of -2.5 to 0.5 dBm and different applied Amplified Stimulated Emission (ASE) to non-linear noise ratios. The total samples were 2160. Simulations and experimental data  for varying optical power in an 800 km link were used and  produced results with a std error of 0.23dB.

In \cite{Wang2019}, a Long Short-Term Memory (LSTM) neural network was used  to approximate the OSNR without need for manual feature extraction. The four tributary output from the coherent receiver was used as input. The LSTM-NN was trained from  simulation of  28/35 GBd PDM 16 and 64-QAM signals and OSNR varied between 15-30 dB. 512 data samples were collected for each OSNR value for a total of 32,768 samples with 70\% used for training and the rest for testing. The Mean Absolute Error (MAE) was found to be 0.1, 0.04, 0.05 and 0.04 dB for 28 GBd PDM 16 and 64-QAM and 35 GBd PDM 16 and 64-QAM respectively. The accuracy of the method was shown to be unaffected by linear impairments of CD and PMD through simulation with variable fiber length. Experimental verification of the model was done on a 34.94 GBd PDM 16-QAM signal with 5,632 samples over an  OSNR range of 15-25 dB, resulting into a MAE of 0.05 dB.
 
The work in  \cite{Khan2017} used a DNN to simultaneously identify modulation format and monitor OSNR. One DNN consisting of 2 hidden layers (45 and 10 neurons respectively) determined the modulation format and then the result was passed to a second stage with multiple 2-hidden layer DNN's (45/40 and 10 neurons respectively) trained per modulation format and the second DNN selected based on 1st stage results.The OSNR could then be predicted for different modulation formats. The input features were obtained from amplitude histograms of varying combinations of modulation formats and OSNR's.  133 experimentally generated AH's for different combinations of modulation format and OSNR  were used to train the DNN's and then tested on 57 AH's  for  112 Gb/s PDM QPSK, 112 Gb/s PDM 16-QAM, and 240 Gb/s PDM 64-QAM signals resulting in mean errors of 1.2, 0.4 and 1 dB respectively. 
\clearpage
%\begin{xltabular}{\textwidth}{p{3cm}p{5cm}Xl}
\begin{table*}
\caption{\bf Performance comparison of existing OPM works for direct detection} \label{tab:6} 
\begin{tabular}{p{3cm}p{5cm}p{7cm}l}
\hline {\textbf{ML algorithm}}  &{\textbf{Features (training:testing)}}&{\textbf{Performance*}}&{\textbf{Ref}} \\\hline 
%\endfirsthead

%\multicolumn{4}{c}%
%{\tablename\ \thetable{} -- continued from previous page} \\
%\hline {\textbf{ML algorithm}}  &{\textbf{Features (training:testing)}}&{\textbf{Performance*}}&{\textbf{Ref}}\\ \hline 
%\endhead

%\hline \multicolumn{4}{|r|}{{Continued on next page}} \\ \hline
%\endfoot

%\hline
%\endlastfoot
SVM& 23 zernike moments from each eye diagram (164:17)&${a=95\% }$ and ${a=60\%\textsuperscript{*} }$ &  \cite{Skoog2006b}\\
ANN(1;12)  &4 inputs from eye diagram  &10 Gbps NRZ-OOK; ${c=0.91}$ \newline 40 Gbps RZ-DPSK; ${c=0.96}$ &\cite{Jargon2009a}\\
ANN & 9600 AH's (80:20) & MAE = 0.167\textsuperscript{*} & \cite{Yuan}\\
ANN(1;28)&7 statistics per ADTP (125:64) &10 Gbps NRZ-OOK; ${c=0.97}$ &\cite{Jargon2009}\\
ANN(1;28)&7 statistics per constellation diagram (216:125) &40 Gbps RZ-QPSK; ${c=0.97}$ \newline RMSE 0.77, 18.71,  1.17 (OSNR,CD,DGD) & \cite{Jargon2010}\\
ANN(1;12)&4 inputs per eye (125:64) \newline \newline \newline \newline (20:12)\textsuperscript{*} \newline \newline (135:32)  & 40 Gbps RZ-OOK; ${c=0.97}$ \newline ME 0.57, 4.68, 1.53 (OSNR, CD, PMD) \newline 40 Gbps RZ-DPSK; ${c=0.96}$ \newline ME 0.77, 4.47, 0.92 (OSNR, CD, PMD) \newline 40 Gbps RZ-OOK; ${c=0.99}$\textsuperscript{*}, \newline ME (0.58, 2.53)\textsuperscript{*} (OSNR, CD) \newline  40 Gbps RZ-DPSK;${c=0.99}$\textsuperscript{*}, ME (1.85, 3.18)\textsuperscript{*} \newline 40 Gbps RZ-DPSK, 3 channel WDM; ${c=0.97}$ \newline ME 0.46, 1.45, 3.98,  0.65 (power, OSNR, CD, PMD) &\cite{Wu2009a}\\
ANN(1;3) & 1 feature per eye (1664:832)\textsuperscript{*} & 32 GBd PDM 64-QAM 0.2dB MSE &\cite{Thrane2017}\\
ANN(1;42)&3627 sets of empirical moments per BR and MF &40 Gbps RZ-DQPSK; RMSE 0.1, 27.3, 0.94 (OSNR, CD, PMD) \newline 56  Gbps RZ-DQPSK; RMSE 0.1, 29, 1.3 \newline 40 Gbps RZ-DPSK; RMSE 0.1, 17, 1 &\cite{Khan2012}\\
Kernel ridge regression & (1200:500) phase portraits, 900 features each (1500:500) \textsuperscript{*}&RMSE +/-11 +/-0.75 (CD, PMD) \newline RMSE +/-11\textsuperscript{*} and +/-1.9\textsuperscript{*}&\cite{Anderson2009}\\
PCA &26,208 ADTP's (70:30; 60:40; 50:50)&10/20 Gbps RZ-OOK, 40/100 Gbps PDM RZ-QPSK,100/200 Gbps PDM NRZ 16-QAM;ME 1, 4, 1.6 (OSNR, CD, PMD) &\cite{Tan2014}\\
ANN(1;12)& (180:144) ADPT's, 7 features each&100 Gbps QPSK ; balanced detection c= 0.995, 0.997* RMSE; 0.45, 1.27* (OSNR) , 3.67, 2.22* (CD), 0.8, 0.91* (PMD) \newline single detection; c=0.96 , RMSE 1.62, 8.75, 7.02 (OSNR, CD, PMD) &\cite{Wu2011}\\
MTL-ANN (1,100;2;50)& (9072:1008) \newline (4320:480)*&28 Gbd NRZ-OOK, PAM4, PAM8; MSE 0.12 \newline 0.11*& \cite{Wan2018}\\
MTDNN-TL(4,100,50,30,2)& (440:243)\textsuperscript{*}  &10 Gbaud PDM 16 , 64 QAM;RMSE 1.09&\cite{Cheng2020}\\
ANN(1;40) & 24 features from PAED &40 Gbps QPSK; ME <20, < 1.3, 1.5-2 (CD, PMD, OSNR) &\cite{Costa2012}\\
MTL-ANN&5 features per OSNR&32 Gbaud PDM 16QAM and 10 Gbaud QPSK MAE 0.28,RMSE 0.48 (OSNR) &\cite{Zheng2020} \\
MTL-CNN&6600 ADTP's&RMSE 0.73, 1.34, 0.47 (OSNR, CD, PMD) &\cite{Fan2018}\\
MTL-CNN&6600 ASCS portraits & RMSE 1.52, 0.81, and 0.32 (Cd, OSNR, PMD) &\cite{Fan2019}\\
MTL-DNN &36,000 AADTPs and AAH's each &MAE 0.2867 (OSNR) a= 99.83\% (CD) &\cite{Luo2021}\\
ANN(5,40,1)&5 statistical features from AH& error <1.1 OSNR&\cite{Zhang2018}\\

\hline
\multicolumn{4}{p{16cm}}{Units for OSNR, PMD, DGD are dB,ps/nm, ps respectively, Performance* indicates experimental results, else simulation results are indicated; a=accuracy, c=corellation}
%\end{xltabular}
\end{tabular}
\label{tab:3}
\end{table*}
\clearpage

This method however was shown to take significant training time and computational power. The same technique was employed in \cite{Yuan} for multiple QAM formats with an added anomaly detector between the MFI ANN and OSNR monitor to improve accuracy. 9600 AH's of 100 bins each were generated for 12.5 GBd signals and 6 modulation formats. The OSNR was varied over the ranges (10-25) for QPSK and 6-QAM, (15-30) for 16-QAM and (20 -35) for 16, 48 and 64-QAM. Experimental results showed a MAE of 0.167 dB.

The authors in \cite{Wang2017} used a CNN to estimate OSNR and recognize modulation format using as input images of constellation diagrams. Simulations were done for 6 modulation techniques i.e. QPSK, 8PSK,  8-QAM, 16-QAM and 32-QAM over OSNR ranges of 15-30 dB  and 64-QAM in the OSNR range of 20-35 dB. Experiments were carried out for 2-QPSK  and 16-QAM. CD was also varied between -100 and 100 ps/nm. The training set consisted of 9600 constellations. The simulation results showed >95\% accuracy for 64-QAM and >99\% accuracy for other formats. They also compared 4 other commonly used algorithms; decision tree with 100 splits, SVM, k-nearest neighbors with 10 neighbors, and BP-ANN with 50 hidden neurons, and found that the CNN achieved better results than the rest at the expense of some computational complexity and large training time. Similar to other methods using constellation diagrams, it performed better for low SNRS <21 dB. Experimental verification was done for QPSK and 16-QAM signals, testing with 20 constellations and results showed maximum error of 0.6 and 0.7 dB respectively.  

In \cite{Xia2019}, a  DNN with transfer learning was studied to monitor OSNR on 56 Gb/s QPSK signals. AH's of the signals were used as input features and trained over an SNR range of 5-35 dB. Each sample AH consisted of 80 bins and the variances were also considered for a total of 81 features per sample. Physical layer parameters were also varied for example launch power (6-8 dB), dispersion (0-600 ps/nm) and bitrates (28-56 Gb/s). The ANN with 5-hidden layer structure bearing 64, 32, 16, 8 and 4 neurons respectively was trained with simulated data and then tested with 128,000 experimentally generated samples, achieving a RMSE of < 0.1 dB.

In \cite{Wang2019b}, four different algorithms were applied  to spectral data from a 20 Gbps QPSK signal i.e. SVM, ANN with 1 hidden layer and 100 hidden neurons, k nearest neighbors with 10 neighbors  and decision tree with 20 splits in a coherent system to estimate OSNR. Training was done with 30 spectra consisting of 4096 samples each collected over an OSNR range between 15-30 dB and the ratio of training:testing data was 2:1. Experimental verification using the same amount of data found that the SVM performed better for the test parameters and took the least computation time. Estimation accuracy was found to be 100, 100, 73.124 and 65.625 for SVM, k-nearest neighbors decision tree and ANN respectively. The poor performance of the ANN was attributed to a large number of input neurons (4096) hence making it prone to under fitting due to increased model complexity. The testing time was also checked and the SVM and KNN found to take the least and longest time respectively.

A binary CNN in which the activation weights were constrained to +/-1 as opposed to floating values was used in \cite{Zhao2020} to predict OSNR for 9 different 12.5 GBd  M-ary QAM signals. Experimental data consisting of gray-scale images of ring constellation diagrams were used. The total dataset consisted of 14,400 images, 100 images per modulation format for each of the 16 OSNR values. With OSNR ranging from 10 - 35 dB, and average accuracy of 98.91\% was found, and was shown to be slightly less accurate than a floating CNN (99.95 \%) and similar to a  multi-layer perceptron (98.86\%) of similar structure, however with reduced energy and execution time. 

In \cite{Yu2019}, the authors used a MTL-ANN to do OSNR estimation and MFI identification similar to their earlier work in \cite{Wan2018}, but applied to a coherent receiver and 9 M-QAM formats at 12.5 GBd. Experimentally generated ring constellation diagrams were transformed to AH's consisting of 200 bins each and used as input features. They were generated over an OSNR range of 10-25 dB for QPSK, 6, 8 and 12-QAM,  15 - 30 dB for 16 and 24-QAM and 20-35 dB for 32, 48, and 64-QAM. 100 AH's were generated per OSNR value and modulation format for a total dataset of 14,400 split into a training:test set of 90:10. The ANN consisted of 1 input layer with 200 neurons, and 2 specific hidden layers for OSNR, while one specific hidden layer was used for MFI, consisting of half the neurons in the previous layer. The optimal neuron number for the shared hidden layer was found to be 350.
Results showed 98.7\% accuracy and RMSE of 0.68 dB when using regression and classification respectively.

A method to simultaneously monitor impairments independent of the signal type was was shown in \cite{Wang2019x}. An LSTM-NN(160,128,2) was used to  predict CD (1360- 2040 ps/nm) and OSNR(15-30 dB) for 28/35 GBd PDM 16/64 QAM signals, using as input the 4 tributary output of the coherent receiver. 512 data samples are generated by simulation for different MF, BR, OSNR and CD and 70\% used for training. The prediction performance obtained was MAE of <0.1dB and 0.64 ps/nm respectively. 

In \cite{Wang2021}, an ANN was shown to estimate OSNR using eigen values consisting of 2nd and 4th order moments and various OSNR's extracted from the rings of the constellation diagrams  as input features. The system was then simulated with 112 Gb/s QPSK, 16 QAM and 120 Gb/s 64-QAM signals and OSNR ranges of 15-26, 19-29 and 22-31 dB respectively. The number of input features for each of the modulation schemes is 3, 3, 9 and the hidden neurons are 5, 5, 12. RMSE's of 0.17, 0.3, 0.68 dB were obtained. Experimental results produced RMSE's of 0.46 and 0.65  for 10/20 Gbd QPSK/16-QAM generated in OSNR ranges of 13-26 and 20-30 dB.

The authors in \cite{feng2020} use a MTL-CNN to experimentally estimate OSNR and identify MF for 28 GBd PDM 8, 16, 32, 64 QAM and 8-PSK and QPSK signals resulting in mean errors of 0.26, 0.4, 0.85, 0.64, 0.17 and 0.19 respectively. A total of 30,600 images of intensity density and differential phase density at different OSNR ranges QPSK (10-30), 8PSK, 8, 16 QAM (12-30), 32 QAM (17-33), 64 QAM (18-33) are used as input features and 85\% used for training. 

The authors in \cite{Ye2021} monitored OSNR  using an LSTM-NN but considered the prediction as a classification problem by defining the continuous OSNR range (15-24 dB) into discrete 1 dB intervals. The NN consisted of 8, 48, 64, 10 neurons for the input, memory, hidden and output layers respectively and the dataset size was 3,000 generated from the IQ output of the coherent receiver, with 75\% of the samples used for training. Simulation was done on a 30 GBd PDM 16 QAM signal resulting in standard deviation within 0.4 dB while experimental verification on a 20 GBd DP-QPSK signal resulted in a standard deviation within 0.67dB.

OPM for few mode fibers was considered in \cite{Saif2021}. In this work, OSNR, CD and mode coupling were monitored with teh aid of  three ML algorithms i.e. SVM, random forest and CNN. The input features were obtained by considering 2D IQH's and their 1D projections in different planes. 200 datasets were generated for each impairment value. In their simulation, the CNN showed the best performance and was then chosen to experimentally verify the accuracy of the proposed technique, resulting in coefficients of determination of 0.98, 0.92 and 0.91 for OSNR, CD and MC respectively. A 10 GBd DP-QPSK signal and ranges of 0-20 dB, 160-1120 ps/nm were used for OSNR and CD respectively, as well as  different mode coupling coefficients.

A single ANN was applied in \cite{Xiang2019} to jointly monitor the MF and OSNR for a 28 GS/s PDM QPSK and 8, 16 and 64 QAM signals over the OSNR range of 10-16, 12-18,15-22 and 22-29 dB respectively. Their ANN  had 50 hidden neurons and took as input two statistical features derived from the amplitude of the signals i.e. kurtosis and variance. Simulation showed mean estimation errors for the OSNR to be 0.005, 0.2, 0.17 and 0.67  using a dataset size of 400 per OSNR and MF. Experimental verification over the ranges 10-17, 14-20, 17-25 dB for QPSK, 8 and 16 QAM showed mean errors 0.15 dB, 0.41 dB and 0.49 dB when 15 hidden neurons are used. The method was extended in \cite{Xiang2021} but 50 bins of the cdf of one stokes parameter was selected as the input. With a a dataset size of 200 per OSNR and MF, OSNR ranges  10-18 dB, 12-20 dB, 12- 20 dB, 16- 24 dB, and 22- 28 dB for QPSK, 8PSK, 8,16, 64 QAM, and 60 hidden neurons, simulation produced mean square errors of 0.086 dB, 0.125 dB, 0.038 dB, 0.17 dB and 0.40 dB. Experimental verification resulted in  mean OSNR estimation error of 0.13 dB, 0.29 dB, and 0.41 dB for QPSK, 8PSK and 16QAM.

Table \ref{table:4} summarizes the current work on OPM for coherent detection.
\subsection{Recognition of Modulation Format}
Many of the OPM methods presented have assumed either advance knowledge of the modulation format or bitrate of the signal, or that it can be obtained from upper layer protocols. As a result, training of the ML algorithms and hence have been investigated for specific modulation formats and bit rates as seen in the previous section and would need to be retrained for a different signal type. It is also not practical to communicate across layers for simple OPM modules \cite{Zhang2016a,Tan2014} therefore it is necessary to review some works which have been done that have identified MFI and/or bitrate.

Since elastic optical networks utilise bandwidth variable transmitters, it would be useful for the OPM module to identify modulation format and bitrate. \cite{Tan2014} proposed one such method using Principal Component Analysis (PCA), where ADTP's for different combinations of bit rate, modulation format and impairments (CD, PMD and OSNR) were generated by simulation and PCA used to create a reference database for the training dataset, and then identified test data with 100\% accuracy in the case when the PC's > 2.

The work in \cite{Khan2017} utilized four DNN's to identify OSNR and MF for three different signal types viz 112 Gbps PM QPSK and 16-QAM and 240 Gbps 64 QAM. One DNN was used to identify the modulation format, and the three DNN's in the second stage trained to estimate the OSNR for one of the three modulation formats. Once the MF was identified, the signal was passed to the respective DNN in stage 2. The method was applied to experimental data from the output of a coherent receiver with AH's used as input features. The method showed 100\% accuracy in all three cases.The authors in \cite{Yuan} proposed an improvement to this method by adding an anomaly detector between the MFI identifier and OSNR monitor to ensure that the MF was accurately identified before being passed to the OSNR monitor. AH's were constructed from constellation diagrams and the method experimentally verified for M-ary QAM. They achieved accuracies of 97.5\%.

A MTL-ANN  in conjuction with signal AH's were applied for MFI and OSNR monitoring in \cite{Wan2018}. Simulation and experiment for NRZ-OOK, PAM 8 and PAM 4 both yielded 100\% accuracy for MFI. The authors extended their work in \cite{Yu2019} to 9 M-QAM modulation formats and used an adaptive weight loss ratio for their ANN as opposed to a fixed optimal one and also achieved 100\% MFI identification accuracy.
ANN's and AAH's were shown to correctly identify six commonly used modulation formats at several datarates and impairment levels with 99.6\% accuracy in \cite{Zhang2016a}.  Similarly, \cite{Huang2021} also used an ANN and AAH's to identify the MF for NRZ, PAM4 and PAM8 signals under stringent bandwidth conditions. The results showed 95\% and 100\% accuracy for simulation and experiment.

Studies were done on the use of a Binary-CNN in \cite{Zhao2020} to identify the MF for 9 different M-ary QAM signals over different OSNR ranges. An experimentally generated data set consisting of 1600 gray scale images of ring constellations per modulation format from the I/Q output of a coherent receiver, with a signal datarate of  12.5 GBd was used. The OSNR was varied from 10-35 dB and all the different formats were identified with 100\% accuracy. This technique  required less memory and execution time compared to a multi-layer perceptron and floating CNN.

In \cite{Zhang2020b}, MFI was done using a CNN that took as input 3 images generated mapping the IQ output from a coherent receiver onto a 3D stokes space, and then projecting it onto 3 2D stokes planes. Numerical simulations were done for 28 GBd PDM signals and 6 modulation formats (BPSK, QPSK, 8,16,32 and 64 QAM) in OSNR conditions varying from 9- 35 dB. 68,400 and 16,200 images in total are used to train and test the CNN respectively. Results show identification accuracy of 99.96\% when the OSNR is above 15 dB.

PCA was used in \cite{Xu2020} to identify the MF of 6 formats (BPSK, QPSK, 8,16,32 and 64 QAM). 3 PC's were extracted from 2048 symbols of the stokes parameters from the received signals of a coherent receiver with OSNR varied from 8 to 40dB and used as a reference database.Testing showed that 100\% MFI accuracy could be obtained at minimum OSNR's of 10, 8, 12, 18, 14 and 23 dB for BPSK, QPSK, 8, 16, 32 and 64 QAM PDM 28 GBd signals respectively. Experimental verification was also done on a dataset containing 30,720 symbols after construction of a reference from 2048 symbols  for 20 GBd QPSK, 8, 16 and 32 QAM signals and also achieved 100\% accuracy.

In \cite{Fan2018} MF and bit rate were determined by a MTL-CNN using 10/20 Gbps RZ-OOK, NRZ-DPSK and NRZ-OOK signals and phase portraits over various impairment ranges for OSNR, CD and PMD. Both MF and BR were identified with 100\% accuracy. 100\% accuracy was also attained by the same authors using a similar MTL-CNN structure but combining features from the different CNN layers and constructing phase portraits from ASCS \cite{Fan2019}.

A multi-input MTL-DNN was used to ascertain the modulation format and bitrate and simultaneously monitor OSNR and CD in \cite{Luo2021}. An experiment was carried out over different OSNR ranges and three CD values using as input AADTPs and AAHs on 14/28 Gbd QPSK and 16QAM signals. MF and BR were identified with accuracy of 100 and 99.81 \% respectively.
\clearpage
\begin{table*}
\caption{\bf Summary of existing OPM works-coherent detection} \label{tab:long} 
\begin{tabular}{p{3cm}p{3.5cm}p{2.8cm}p{3cm}p{3cm}l}
\hline 
%\multicolumn{1}{l}{\textbf{ML algorithm}} & \multicolumn{1}{l}{\textbf{Signal type (BR-MF)}} & \multicolumn{1}{l}{\textbf{Input features}} &\multicolumn{1}{l}{\textbf{Impairment}}&\multicolumn{1}{l}{\textbf{Performance}}&\multicolumn{1}{l}{\textbf{Ref}}\\ \hline 
{\textbf{ML algorithm}} & {\textbf{Signal type (BR-MF)}} & {\textbf{Input features}} &{\textbf{Impairment}}&{\textbf{Performance}}&{\textbf{Ref}}\\ \hline 
%\endfirsthead

%\multicolumn{5}{c}%
%{\tablename\ \thetable{} -- continued from previous page} \\
%\hline \multicolumn{1}{c}{\textbf{ML algorithm}} & \multicolumn{1}{c}{\textbf{Signal type}} & \multicolumn{1}{c}{\textbf{Input features}} &\multicolumn{1}{c}{\textbf{Impairment}}&\multicolumn{1}{c}{\textbf{Performance}}&\multicolumn{1}{c}{\textbf{Ref}}\\ \hline 
%\endhead

%\hline \multicolumn{6}{|r|}{{Continued on next page}} \\ \hline
%\endfoot

%\hline
%\endlastfoot

DNN(5;500)&16\textsuperscript{b}- QPSK&IQ output &OSNR (7.5-31)&mean error 1.6 &\cite{Tanimura2016}\\
CNN&14\textsuperscript{b}- and 16\textsuperscript{b}- QPSK, 16 QAM, 64 QAM&IQ output &OSNR (11-33)&bias error <0.2&\cite{Cho2019}\\
ANN(1;6)&56.8\textsuperscript{b}-16 QAM&link parameters, ANC &OSNR&error <0.6&\cite{Kashi2017}\\
ANN(1;7)& 35\textsuperscript{b}-16 QAM&link parameters, ANC &non-linear SNR&std error <0.23& \cite{Caballero2018a}\\
ANN& 12.5\textsuperscript{b}- M-ary QAM & AH &OSNR (10-35)&MAE 0.167 \\
LSTM NN&28\textsuperscript{b}-/35\textsuperscript{b}-16 and 64 QAM&IQ output&OSNR(15 - 30)&MAE 0.1/0.05, 0.04/0.04&\cite{Wang2019}\\
DNN(2;45/10)& 112- QPSK, 16-QAM, and 240- 64 QAM&AH&OSNR&mean errors 1.2, 0.4, 1&\cite{Khan2017}\\
CNN& 25\textsuperscript{b}- QPSK, 8PSK, 8 QAM, 16 QAM, 32 QAM %\newline 
25\textsuperscript{b}- 64 QAM,%\newline
25\textsuperscript{b}- QPSK\textsuperscript{*}, 16 QAM\textsuperscript{*} &constellation &OSNR(15-30) %\newline \newline \newline
OSNR(20-35)&>99\%accuracy %\newline \newline \newline
> 95\% %\newline
max. error 0.6\textsuperscript{*},  0.7\textsuperscript{*}&\cite{Wang2017}\\
TL-DNN&56-  QPSK&AH&OSNR(6-20)&RMSE    <0.1&\cite{Xia2019}\\
SVM, ANN(1;100), K nearest neighbors, Decision tree  &20- QPSK&Spectrum&OSNR& accuracy 100\% , 65.625\%, 100\%, 73.124\%& \cite{Wang2019b}\\
CNN&12.5\textsuperscript{b}- M-ary QAM &ring constellation &OSNR (10-35)&98.91 accuracy &\cite{Zhao2020}\\
MTL-ANN&12.5\textsuperscript{b} M-QAM&AH&OSNR(10-35)&accuracy 98.7\%&\cite{Yu2019}\\
LSTM-NN&28\textsuperscript{b}- /35\textsuperscript{b}- 16/64 QAM&IQ output&OSNR(15-30), CD(1360-2040)&MAE <0.1 and <0.64&\cite{Wang2019x}\\
ANN&112- QPSK, 16 QAM , 120- 64 QAM%newline
10\textsuperscript{b}/20\textsuperscript{b}- QPSK\textsuperscript{*} 16 QAM\textsuperscript{*} & ring constellations& OSNR(15-26,19-29,22-31) %\newline
OSNR(13-26, 20-30)\textsuperscript{*} &RMSE 0.17, 0.3, 0.68 %\newline  \newline 
RMSE 0.46 and 0.65* &\cite{Wang2021}\\
MTL-CNN& 28\textsuperscript{b}- (8, 16), 32, 64 QAM and 8-PSK, QPSK& intensity and differential phase density diagrams &OSNR (12-30), (17-33), (18-33), (12-30),  (10-30)&mean errors 0.26, 0.4, 0.85, 0.64, 0.17, 0.19&\cite{feng2020}\\
LSTM-NN& 30\textsuperscript{b}- 16 QAM, 30 \textsuperscript{b}- QPSK\textsuperscript{*}&IQ output &OSNR(15-24)&STD <0.4 , <0.67\textsuperscript{*}&\cite{Ye2021}\\
CNN&10\textsuperscript{b}- QPSK &IQH & OSNR (0-20), CD (160-1120) and different mode coupling coefficients & coefficients 0.98, 0.92, 0.91 &\cite{Saif2021}\\
ANN&28- QPSK, 8, 16, 64 QAM %\newline
QPSK, 8 and 16 QAM\textsuperscript{*}& statistics from IQ & OSNR (10-16, 12-18, 15-22, 22-29) %\newline 
(10-17, 14-20, 17-25)\textsuperscript{*}&mean errors 0.005, 0.2, 0.17, 0.67 %\newline 
(0.15, 0.41, 0.49)\textsuperscript{*}&\cite{Xiang2019}\\
ANN&28- QPSK, 8 PSK, 8, 16, 64 QAM %\newline \newline
QPSK, 8PSK, 16 QAM\textsuperscript{*}& Stokes parameters & OSNR (10-18, 12-20, 12- 20, 16- 24, and 22- 28) %\newline 
(9.8-16.8, 12-19,  16-23)\textsuperscript{*}&MSE 0.086, 0.125, 0.038, 0.17, 0.40. Mean error (0.13, 0.29,0.41) \textsuperscript{*}&\cite{Xiang2021}\\
\hline
\multicolumn{5}{p{18cm}}{All units for OSNR,CD,PMD are in dB,ps/nm and ps respectively. *indicates experimental values and simulation otherwise. BR\textsuperscript{b}-MF represents bitrate(GBd)-modulation format and bitrate in Gbps otherwise. ANN(x;y): x and y represent the number of hidden layers and neurons in the hidden layer respectively.}
\end{tabular}
\end{table*}
\clearpage
\label{table:4}
%\end{wraptable}

In \cite{Fan2018} MF and bit rate were identified by a MTL-CNN using 10/20 Gbps RZ-OOK, NRZ-DPSK and NRZ-OOK signals and phase portraits over various impairment ranges for OSNR, CD and PMD. Both MF and BR were identified with 100\% accuracy. 100\% accuracy was also attained by the same authors using a similar MTL-CNN structure but combining features from the different CNN layers and constructing phase portraits from ASCS \cite{Fan2019}.

A multi-input MTL-DNN was used to find modulation format and bitrate and simultaneously monitor OSNR and CD in \cite{Luo2021}. An experiment was carried out over different OSNR ranges and three CD values using as input AADTPs and AAHs on 14/28 Gbd QPSK and 16QAM signals. MF and BR were identified with accuracy of 100 and 99.81 \% respectively. 

The authors in \cite{Zhang2018} used a 3-layer ANN(202,40,5) to identify 5 QAM formats in an experimental IM-DD QAM-OFDM system using AH's as input. The MFI accuracy obtained was close to 100\% for 4 and 16 QAM over the entire range of received optical power, while 32, 64 and 128 QAM got similar accuracy when the optical powerexceededd -11dBm. 

In \cite{feng2020} a MTL-CNN was shown to identify MF with 100\% accuracy for mPSK and mQAM signals at a baud rate of 28 GBd and OSNR varied from 10-33dB.

A 3-layer ANN was also shown in \cite{Xiang2019,Xiang2021} that achieved 100\% MFI accuracy for different values of OSNR between 10-28 dB for 5 modulation formats.

Table \ref{table:5} summarizes some works where MFR has been done.
\begin{table*}
 \centering
  \captionsetup[]{justification=centering}
  \caption{\bf Summary of ML methods used for MFI}
\begin{tabular}{lp{2.5cm}p{5cm}p{2.5cm}l}
\hline
\textbf{ML method} & \textbf{Feature type} & \textbf{Modulation Format} & \textbf{Accuracy (\%)}&\textbf{Reference} \\ \hline
PCA&ADTP&10/20 Gb/s RZ-OOK, 40/100 Gb/s PM RZ QPSK, 100/200 Gb/s PM NRZ 16QAM & 100& \cite{Tan2014}\\ 
DNN & AH & 112 Gb/s PM QPSK ,PM-16QAM, 240 Gb/s-PM 64-QAM & 97.5 &\cite{Khan2017}\\
ANN&AH& M-ary QAM&95.7&\cite{Yuan}\\
MTL-ANN&AH&NRZ-OOK, PAM4 , PAM8 & 100&\cite{Wan2018} \\
&&M-QAM&100&\cite{Yu2019}\\
ANN&AH&10 Gb/s RZ-OOK, 40 Gb/s NRZ-DPSK, 40 Gb/s ODB, 40 Gb/s RZ-DQPSK, 100 Gb/s PM RZ-QPSK, 200 Gb/s PM-NRZ 16QAM &99.6&\cite{Zhang2016a}\\
ANN&AH&NRZ, PAM4, PAM8&95 simulation,\newline 100 experiment &\cite{Huang2021}\\
B-CNN&ring constellation images&M-ary QAM&100 experiment & \cite{Zhao2020}\\
CNN&2D stokes plane images &M-QAM &99.96&\cite{Zhang2020b}\\
PCA&Stokes parameters&M-QAM&100&\cite{Xu2020}\\
MTL-CNN&ADTPs&10/20Gb/s RZ-OOK, NRZ-OOK, NRZ-DPSK &100 &\cite{Fan2018}\\
MTL-ANN&ASCS phase portraits & 60/100 Gb/s QPSK, 16, 64 QAM&100&\cite{Fan2019}\\
MTL-DNN &AADTP's and AAH's&14/28 Gbd QPSK, 16 QAM & 100 & \cite{Luo2021}\\
ANN(202,40,5)&AHs&4,16,32,64,128 QAM&close to 100 &\cite{Zhang2018}\\
MTL-CNN&intensity density and differential phase density diagram & 28 GBd mQAM and mPSK &100&\cite{feng2020}\\
ANN&amplitude statistics and stokes parameter&mPSK and mQAM &100&\cite{Xiang2019,Xiang2021}\\
\end{tabular}
\label{table:5}
\end{table*}
\subsection{Application of Photonic Reservoir Computing in OPM}

Photonic reservoir computing in the optical domain has been considered as an alternative to Digital Signal Processing for some years\cite{Pachnike2020}. A reservoir computer (RC) typically consists of an input, reservoir and readout. A input signal is fed to the reservoir, consisting of multiple randomly connected non-linear nodes, that function like a neural network. The input signal can alter the current and future states of the reservoir. The output of the reservoir is then readout as a linear combination of the different states in the reservoir. The input weights and node connections are fixed and thus the training complexity is reduced to a linear one at a single node at the readout \cite{Pachnike2020,Vandoorne2008,appeltant2011}. A common implementation that has been presented in the literature uses a single non-linear element in combination with a delay loop \cite{appeltant2011}, which can be implemented in the optical domain using a semiconductor laser and a fiber loop \cite{appeltant2011,Brunner2013,Larger2012}. Other approaches have used a network of several interconnected Semiconductor Optical Amplifiers(SOA's) \cite{Vandoorne2008,VANDOORNE2011}, and  silicon micro-ring resonators \cite{Mesaritakis2013}. \cite{Vandoorne2014} has also shown a RC implementation using a passive silicon chip where the non-linearity is transferred to the readout, whose output is then passed to a linear classifier. 
Implementing the RC using photonic devices brings  several advantages such as speed due to their inherently parallel computation nature, low power consumption  and high bandwidth operation which are direct results of using light rather than electrical signals \cite{Larger2012,Vandoorne2014}.
The authors in \cite{Ai2021} have applied this concept of reservoir computing using a semiconductor laser and delay line to  identify the modulation format of 10 Gb/s OOK, 40 Gb/s DQPSK and 100 Gb/s 16-QAM signals in varying OSNR (12-26 dB), CD (-500 to 500 ps/nm) and DGD (0-20 ps) conditions. The input features were derived from AAH's. From a dataset size of 11,700, 2700 modulation signals were used to train the the model using ridge regression and 100 samples used for testing. The training and testing process is repeated five times with the different sample sets and using 400 virtual nodes. The method achieved a classification accuracy of 95.1\%, 95.7\% and 95.5\% for OOK, DPQSK and 16-QAM.
\section{Discussion}
The most common features used in the current OPM works for feature selection are eye diagrams, phase portraits and amplitude histograms. In some cases, widely known features from these plots such as statistical means, variances, standard deviations etc, counts of occurrences per bin, eye diagram parameters like eye closure, crossing amplitude etc have been used, while in others new features have been defined to exploit visible differences in the plots \cite{Jargon2009,Caballero2018,Saif2021}. Manual definition of features is a difficult task which requires experience and also makes it impossible to distinguish patterns when there are only slight differences for example, the performance of ANN's have been shown to deteriorate beyond certain OSNR's because there is very little distinction between the eye diagrams especially for higher modulation formats\cite{Thrane2017}. It also makes it difficult to scale the ML algorithm to a different signal type than what it was trained with. To mitigate this, deep learning techniques have been studied where the algorithm can learn its own features from the input data, the commonest way being by supplying it with processed images\cite{Fan2019,Fan2020,Wang2017,Zhang2020b} and the 4-tributary output of the coherent receiver. Of course, this comes with more complexity since deep learning algorithms are generally more difficult to train. Furthermore, in cases where images are used, some amount of image processing is required \cite{Zhang2020b,Skoog2006b}.

Artificial neural networks have been very widely used for OPM in direct detection systems. The reviewed works have shown that in some cases, even simple ANN's with 1 hidden layer and as low as 3 hidden neurons and as few as one input feature are capable of accurately predicting OSNR, CD and PMD. Correlations of upto 0.997 have been obtained. The performance of the ANN depends on the input features selected and their number and also on the signal type. SVMs, PCA and ridge regression have also been used for but in very limited works. Deep learning techniques have also been shown in the literature but require significant time and more features to accurately train. 

Many of the techniques used are dependent on the signal type hence it is assumed that the monitoring unit already has knowledge of the signal type. Moreover, in the cases where multi-impairment monitoring is required of different signal types, the ANN has to be trained more than once or multiple ANN's have to be used for each signal type. \cite{Tan2014} proposed a method using PCA  and that was transparent to the BR and MF but required training with multiple combinations of MF-BR-impairments hence required a significant amount of training data. More recently, \cite{Zheng2020} has shown a method which is transparent to the signal type and only requires input power as a feature. However, it has only been used to measure OSNR. Other works have also utilised multi-task learning and deep learning \cite{Cheng2020,Fan2018,Wan2018} to simultaneously identify the signal type and impairments. These also required generating large training datasets with different combinations of the signal type and impairment levels. 
Very few works have measured other impairments such as non-linearity whose monitoring is also crucial for optical networks.

For coherent detection systems, neural networks have been used and shown to perform better than other methods where there have been compared except in one case in \cite{Wang2019b}. ANN's still suffer from manual feature generation and as such most of the literature uses DNN's and CNN's for coherent detection systems which can learn their own features from the 4 tributary output of the coherent receiver, images of constellations in the Jones or Stokes space or AH's. The challenge is that the training takes a considerable amount of time and a very large number of samples are required to produce accurate models. Nevertheless, after the training stage, the monitoring stage takes a shorter time, which is the critical time for an OPM monitor in a real system, since he training can be done off-line. Many of the methods have also been shown to maintain their accuracy in the presence of linear impairments. 
\cite{Zhao2020} tried to compare the performance of their joint MFI and SNR predictor by simulation for different transmission parameters noting that future networks will have varying parameters. They varied the transmission distance and launch power. They showed that if the DNN's were trained each time there was a change in one parameter, 100\% accuracy could be obtained for both MFI and OPM, whereas lower accuracy was obtained if trained once with a dataset consisting of all the possible parameter variations.

It is difficult to directly compare one ML implementation in one work over the other because different authors have carried out their simulations/experiments for different impairment ranges, signal types and they have classified the performance of their algorithms in different ways. 

In the reviewed literature where MFR and BRI have been investigated, again ANN's and deep learning neural networks have been the most common method of choice and the bulk of the work has achieved 100\% identification accuracy.

Photonic reservoir computing is a promising technology for OPM and MFR since it reduces the training complexity of neural network based methods which has been highighted as a key challenge in teh reviewed works that have employed them. Moroever, signal processing in the optical domain allows for high speed and high bandwidtch operation which are critical for future communication networks. 

\section{Conclusion}
Optical performance monitoring has been an important aspect of optical communications for a very long time. As networks have become more heterogeneous and dynamic, they have also become more complex and fiber network technology has had to evolve along with it to meet the reliability demands, since it can already provide the required capacity. Since the light paths are expected to constantly change as they become elastic to provide bandwidth on demand, and the signal parameters are also expected to change during transmission in accordance with link conditions, real time link performance has become important. Application of machine learning to OPM has garnered  significant interest as a promising technology to aid in this task and has been shown to be possible, and to provide accurate prediction for multiple impairments as long as the algorithm is well trained.  

%\section{Acknowledgments}
\bigskip

\section{Disclosures}
\noindent\textbf{Disclosures.} The authors declare no conflicts of interest.

%\section{References}

\bigskip

% Bibliography
\bibliography{Optical.bib}
% Full bibliography added automatically for Optics Letters submissions; the following line will simply be ignored if submitting to other journals.
% Note that this extra page will not count against page length
\bibliographyfullrefs{sample}
%Manual citation list
%\begin{thebibliography}{1}
%\bibitem{Zhang:14}
%Y.~Zhang, S.~Qiao, L.~Sun, Q.~W. Shi, W.~Huang, %L.~Li, and Z.~Yang,
 % \enquote{Photoinduced active terahertz metamaterials with nanostructured
  %vanadium dioxide film deposited by sol-gel method,} Opt. Express \textbf{22},
  %11070--11078 (2014).
%\end{thebibliography}
% Please include bios and photos of all authors for aop articles
\ifthenelse{\equal{\journalref}{aop}}{%
\section*{Author Biographies}
\begingroup
\setlength\intextsep{0pt}
\begin{minipage}[t][6.3cm][t]{1.0\textwidth} % Adjust height [6.3cm] as required for separation of bio photos.
  \begin{wrapfigure}{L}{0.25\textwidth}
    \includegraphics[width=0.25\textwidth]{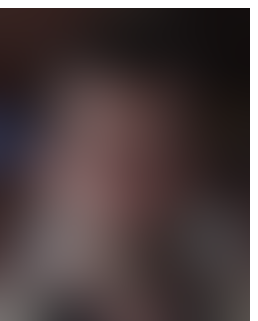}
  \end{wrapfigure}
  \noindent
  {\bfseries John Smith} received his BSc (Mathematics) in 2000 from The University of Maryland. His research interests include lasers and optics.
\end{minipage}
\begin{minipage}{1.0\textwidth}
  \begin{wrapfigure}{L}{0.25\textwidth}
    \includegraphics[width=0.25\textwidth]{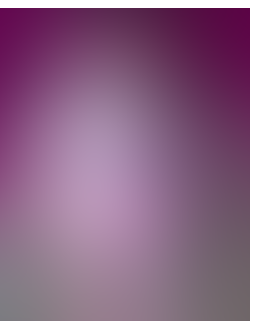}
  \end{wrapfigure}
  \noindent
  {\bfseries Alice Smith} also received her BSc (Mathematics) in 2000 from The University of Maryland. Her research interests also include lasers and optics.
\end{minipage}
\endgroup
}{}

\end{document}